# A new Parameter Learning Method for Bayesian Networks with Qualitative Influences

Ad Feelders
Department of Information and Computing Sciences, Utrecht University
PO Box 80089, 3508 TB Utrecht, The Netherlands

## Abstract

We propose a new method for parameter learning in Bayesian networks with qualitative influences. This method extends our previous work from networks of binary variables to networks of discrete variables with ordered values. The specified qualitative influences correspond to certain order restrictions on the parameters in the network. These parameters may therefore be estimated using constrained maximum likelihood estimation. We propose an alternative method, based on the isotonic regression. The constrained maximum likelihood estimates are fairly complicated to compute, whereas computation of the isotonic regression estimates only requires the repeated application of the Pool Adjacent Violators algorithm for linear orders. Therefore, the isotonic regression estimator is to be preferred from the viewpoint of computational complexity. Through experiments on simulated and real data, we show that the new learning method is competitive in performance to the constrained maximum likelihood estimator, and that both estimators improve on the standard estimator.

## 1 Introduction

In recent work, Wittig and Jameson (2000), Altendorf, Restificar, and Dietterich (2005) and Feelders and van der Gaag (2005) have shown that the use of qualitative influences can improve the probability estimates in Bayesian networks, in case relatively few observations are available. Feelders and van der Gaag (2005) provide a simple algorithm, based on the isotonic regression, that is however restricted to binary variables. On the other hand, the approaches of Wittig and Jameson (2000), and Altendorf, Restificar, and Dietterich (2005) can handle non-binary variables as well, but require a numeric optimization algorithm to compute the estimates. Apart from improvement of the parameter estimates, Wittig and Jameson (2000) and Feelders and van der Gaag (2005) argue that networks with probability estimates that reflect the qualitative knowledge specified by the domain experts are less likely to exhibit counterintuitive reasoning behaviour and are therefore more likely to be accepted.

In this paper, we propose an alternative order-restricted estimator for discrete variables with ordered values. This estimator is easy to compute, and, as our experiments show, performs quite well compared to the order-restricted maximum likelihood estimator. We note that the work discussed in Niculescu, Mitchell, and Rao (2006) and Niculescu, Mitchell, and Rao (2007) on Bayesian network learning with parameter constraints, does not cover the type of parameter constraints resulting from qualitative influences, and can therefore be seen as complementary to the work presented here.

The paper is organized as follows. In section 2 we introduce the required notation that will be used throughout the paper. In section 3 we describe the proposed parameter learning method. In section 4 we discuss an algorithm for order-restricted maximum likelihood estimation due to Hoff (2003). This algorithm is of interest in its own right, and is used in our experiments for comparison with the proposed alternative estimator. Section 5 reports on experiments with real and artificial data sets. Finally, section 6 concludes.

## 2 Preliminaries

We consider a node $Y$, with parents $\mathbf{X} = X_1, \ldots, X_k$. Let $\mathcal{Y} = \{1, 2, \ldots, d_y\}$ denote the set of $d_y$ possible values of $Y$. We assume the values of $Y$ are linearly ordered, with the obvious ordering. Likewise, $\mathcal{X}_i = \{1, 2, \ldots, d_i\}$ denotes the set of $d_i$ possible values of $X_i$. The values of $X_i$ are again assumed to be linearly ordered, with the obvious ordering. Furthermore, let $\mathcal{X} = \times_{i=1}^{k} \mathcal{X}_i$ denote the set of all parent configurations of $Y$.

A parent $X_i$ has a *positive* influence on $Y$ ($X_i \xrightarrow{+} Y$) if for all $x_i, x_i' \in \mathcal{X}_i$:

$$x_i \leq x_i' \Rightarrow P(Y|x_i, \mathbf{x}_{-i}) \preceq P(Y|x_i', \mathbf{x}_{-i}) \quad (1)$$

where $\mathbf{x}_{-i}$ is any configuration of the parents of $Y$ other than $X_i$ (Wellman 1990). Here $P(Y|x_i, \mathbf{x}_{-i}) \preceq P(Y|x_i', \mathbf{x}_{-i})$ means that the distribution of $Y$ for parent configuration $(x_i, \mathbf{x}_{-i})$ is stochastically smaller than for parent configuration $(x_i', \mathbf{x}_{-i})$, that is

$$F(y|x_i, \mathbf{x}_{-i}) \geq F(y|x_i', \mathbf{x}_{-i}), \quad y = 1, 2, \ldots, d_y$$

where $F(y) = P(Y \leq y)$. In words: for the smaller value of $X_i$, smaller values of $Y$ are more likely. A negative qualitative influence is defined analogously, where for larger values of $X_i$ smaller values of $Y$ are more likely. Without loss of generality, we henceforth assume that all influences are positive. A negative influence from $X_i$ to $Y$ can be made positive simply by reordering the values in $\mathcal{X}_i$.

In the sequel, it will be convenient to impose the ordering specified in (1) on the parent configurations, together with the rule that

$$\forall \mathbf{x}, \mathbf{x}' \in \mathcal{X} : \mathbf{x} \preceq \mathbf{x}' \Rightarrow P(Y|\mathbf{x}) \preceq P(Y|\mathbf{x}'),$$

where the order on $\mathcal{X}$ is the product order

$$\mathbf{x} \preceq \mathbf{x}' \equiv \forall i = 1, \ldots, k : x_i \leq x_i'$$

## 3 The isotonic regression estimator

Before we start the discussion of our proposed estimator, we note that qualitative influences only impose constraints between the parameters (conditional probabilities) associated with the different parent configurations of a variable, but the parameters associated with different variables are still unrelated. Hence, we can in the sequel restrict our attention to the parameters associated with a single variable.

Recall that the qualitative influences impose a partial order $\preceq$ on $\mathcal{X}$. Any real-valued function $f$ on $\mathcal{X}$ is *isotonic* with respect to $\preceq$ if, for any $\mathbf{x}, \mathbf{x}' \in \mathcal{X}$, $\mathbf{x} \preceq \mathbf{x}'$ implies $f(\mathbf{x}) \leq f(\mathbf{x}')$. Likewise, any real-valued function $f$ is *antitonic* with respect to $\preceq$ if, for any $\mathbf{x}, \mathbf{x}' \in \mathcal{X}$, $\mathbf{x} \preceq \mathbf{x}'$ implies $f(\mathbf{x}) \geq f(\mathbf{x}')$. Given real-valued numbers $g(\mathbf{x})$ and positive weights $w(\mathbf{x})$, a function $g^*$ is an *isotonic (antitonic) regression* of $g$ with respect to the weight function $w$ and partial order $\preceq$, if and only if it minimizes the sum

$$\sum_{\mathbf{x} \in \mathcal{X}} w(\mathbf{x}) \left[ f(\mathbf{x}) - g(\mathbf{x}) \right]^2$$

within the class of isotonic (antitonic) functions $f$ on $\mathcal{X}$. The existence of a unique $g^*$ has been proven by Brunk (1965).

Let $\hat{F}(y|\mathbf{x}) = \sum_{y' \leq y} \hat{P}(y'|\mathbf{x})$ denote the unconstrained MLE of $F(y|\mathbf{x}), \mathbf{x} \in \mathcal{X}$, and let $n(\mathbf{x})$ denote the number of observations with parent configuration $\mathbf{x}$.

**Definition 1 (Isotonic regression estimator)**
*The isotonic regression estimator*

$$\hat{F}^*(y|\mathbf{x}), \qquad y = 1, 2, \ldots, d_y; \quad \mathbf{x} \in \mathcal{X}$$

*for partial stochastic order, is given by the antitonic regression of $g(\mathbf{x}) = \hat{F}(y|\mathbf{x})$ with weights $w(\mathbf{x}) = n(\mathbf{x})$, for each value $y = 1, 2, \ldots, d_y$.*

This estimator is an extension of one proposed for linear orders already by Hogg (1965), and later analyzed by El Barmi and Mukerjee (2005).

It is not hard to verify that this estimator satisfies the order constraints by construction. Also, it is an intuitively plausible estimator, since it returns the estimates with minimum weighted squared distance to the observed cumulative relative frequencies, that satisfy the constraints. For binary variable $Y$, the proposed estimator coincides with the constrained maximum likelihood estimator, but this is no longer the case if $Y$ can take on more than two values.

One should verify whether $\hat{F}^*(Y|\mathbf{x})$ is indeed a valid distribution function, as required. Hence, we have to check two properties:

1. $\forall y \in \mathcal{Y} : 0 \leq \hat{F}^*(y|\mathbf{x}) \leq 1$.

2. $y \geq y' \Rightarrow \hat{F}^*(y|\mathbf{x}) \geq \hat{F}^*(y'|\mathbf{x})$.

Both follow almost immediately from well-known properties of the isotonic regression. To show this, we require some additional definitions. A subset $L$ of $\mathcal{X}$ is a *lower set* of $\mathcal{X}$ with respect to $\preceq$, if $\mathbf{x} \in L$, $\mathbf{x}' \in \mathcal{X}$, and $\mathbf{x}' \preceq \mathbf{x}$ imply $\mathbf{x}' \in L$. Hence, if a lower set contains a particular element, it is required to also contain all lower ordered elements. Likewise, a subset $U$ of $\mathcal{X}$ is an *upper set* of $\mathcal{X}$ if $\mathbf{x} \in U$, $\mathbf{x}' \in \mathcal{X}$, and $\mathbf{x} \preceq \mathbf{x}'$ imply $\mathbf{x}' \in U$. The *weighted average* of $g$, with weights $w$, for a nonempty subset $A$ of $\mathcal{X}$ is defined as

$$\mathrm{Av}(A) = \frac{\sum_{\mathbf{x} \in A} w(\mathbf{x}) g(\mathbf{x})}{\sum_{\mathbf{x} \in A} w(\mathbf{x})}$$

Now, the antitonic regression can be characterized as follows (Robertson, Wright, and Dykstra 1988):

$$g^*(\mathbf{x}) = \max_{L:\mathbf{x} \in L} \min_{U:\mathbf{x} \in U} \mathrm{Av}(L \cap U) \quad (2)$$

From (2) we can see that $\hat{F}^*(y|\mathbf{x})$ is a weighted average of observed relative frequencies, so it follows that $0 \leq \hat{F}^*(y|\mathbf{x}) \leq 1$. Also, since we have that $\hat{F}(y|\mathbf{x}) \geq \hat{F}(y'|\mathbf{x})$ for $y \geq y'$, it follows that its weighted average is larger for every set $A$. Hence, it follows from (2) that $\hat{F}^*(y|\mathbf{x}) \geq \hat{F}^*(y'|\mathbf{x})$, as required.

Next, we turn to the computation of the estimator. Let $N$ denote the number of elements of $\mathcal{X}$. The best time complexity known for an exact solution to the isotonic regression problem for arbitrary partial order on $\mathcal{X}$ is $O(N^4)$ (Maxwell and Muchstadt 1985), so the time complexity for computing the proposed estimator using this algorithm is $O(d_y N^4)$. Dykstra and Robertson (1982), however, propose an efficient iterative algorithm for product partial orders. This algorithm is very simple to implement, because it only requires the repeated application of the Pool Adjacent Violators (PAV) algorithm for linear orders. The PAV algorithm runs in linear time (Ahuja and Orlin 2001).

The pseudo-code for the product order algorithm of Dykstra and Robertson (1982) is given below. It takes as input numbers $g(\mathbf{x})$ and weights $w(\mathbf{x})$, $\mathbf{x} \in \mathcal{X}$ to be made isotonic with respect to the product order on $\mathcal{X}$.

**IRProductOrder**($g(\mathbf{x}), w(\mathbf{x})$)
  **For** each $j \in \{1, \ldots, k\}$ and $\mathbf{x} \in \mathcal{X}$ **do**
    $I_j(\mathbf{x}) \leftarrow 0$
  **Repeat**
    **For** each $j \in \{1, \ldots, k\}$ **do**
      **For** each $\mathbf{x} \in \mathcal{X}$ **do**
        $g_j(\mathbf{x}) \leftarrow g(\mathbf{x}) + \sum_{i \neq j} I_i(\mathbf{x})$
      **For** each $\mathbf{x}_{-j} \in \mathcal{X}_{-j}$ **do**
        $\hat{g}_j(\mathbf{x}_{-j}, X_j = 1), \ldots, \hat{g}_j(\mathbf{x}_{-j}, X_j = d_j) \leftarrow$
        PAV($g_j(\mathbf{x}_{-j}, X_j = 1), \ldots, g_j(\mathbf{x}_{-j}, X_j = d_j)$,
        $w(\mathbf{x}_{-j}, X_j = 1), \ldots, w(\mathbf{x}_{-j}, X_j = d_j)$))
      **For** each $\mathbf{x} \in \mathcal{X}$ **do**
        $I_j(\mathbf{x}) \leftarrow \hat{g}_j(\mathbf{x}) - g_j(\mathbf{x})$
  **Until** converged
  **Return** $\hat{g}(\mathbf{x})$

For each parent $X_j$, and each configuration $\mathbf{x}_{-j}$ of the remaining parents the algorithm performs the isotonic regression on the vector

$$g_j(\mathbf{x}_{-j}, X_j = 1), \ldots, g_j(\mathbf{x}_{-j}, X_j = d_j),$$

where $(\mathbf{x}_{-j}, X_j = x_j)$ denotes the parent configuration with $\mathbf{X}_{-j} = \mathbf{x}_{-j}$ and $X_j = x_j$. This set of parent configurations has linear order, so the PAV algorithm can be applied to make the vector decreasing (recall we require the antitonic rather than the isotonic regression). The numbers $g_j(\mathbf{x})$ are obtained by adding the "increments" $I_i(\mathbf{x}), i \neq j$ from the other parents to $g(\mathbf{x})$. Dykstra and Robertson (1982) prove that the $\hat{g}_j, j = 1, 2, \ldots, k$ all converge to the isotonic regression $g^*$. To compute the parameter estimates, this algorithm is called with $g(\mathbf{x}) = \hat{F}(y|\mathbf{x})$ and $w(\mathbf{x}) = n(\mathbf{x})$ for each value of $Y$.

Since there are $k$ parents, and the number of values of parent $j$ is $d_j$, the number of operations within the repeat loop can be computed as follows. To process parent $j$ we have to perform $\prod_{i \neq j} d_i$ isotonic regressions on $d_j$ distinct values. Each isotonic regression can be computed in $O(d_j)$ steps, so to process parent $j$ we have to perform $O(N)$ operations. Since we have to do this for each parent, each repeat cycle takes $O(kN)$ operations.

## 4 Maximum Likelihood Estimation

The constrained maximum likelihood estimates are obtained by maximizing

$$\mathcal{L} = \sum_{\mathbf{x} \in \mathcal{X}} \sum_{y \in \mathcal{Y}} n(y, \mathbf{x}) \log P(y|\mathbf{x})$$

subject to: $\mathbf{x} \preceq \mathbf{x}' \Rightarrow P(Y|\mathbf{x}) \preceq P(Y|\mathbf{x}')$, and where $n(y, \mathbf{x})$ denotes the number of cases with parent configuration $\mathbf{x}$ and $Y = y$.

Altendorf, Restificar, and Dietterich (2005) translate the order constraints into penalty terms in the objective function, and use a BFGS-type algorithm to optimize the penalized log-likelihood function. Their algorithm may converge outside the feasible region, in which case the penalty weight is increased and the optimization is performed anew. They mention more efficient and reliable optimization methods as a point for further research. Dykstra and Feltz (1989) give an algorithm for computing the constrained maximum likelihood estimates, but this algorithm is quite complicated. For running our experiments, we have chosen an algorithm proposed by Hoff (2003). This algorithm is very easy to implement, and will produce the constrained maximum likelihood estimates like the algorithm of Dykstra and Feltz (1989).

The algorithm of Hoff (2003) is based on the following observations. Recall that the goal is to estimate a *collection* of distributions (one for each parent configuration) subject to stochastic order constraints. The set of all collections satisfying these constraints is a convex set, that is, if we take a convex combination of any two collections satisfying the constraints, then the resulting collection will satisfy the constraints as well. Note furthermore that each element of a convex set can be written as a convex combination of its extreme elements. As Hoff (2003) shows, an extreme element corresponds to a collection of point-mass distributions that satisfies the constraints. From these observations it follows that we can view the estimation problem currently under consideration, as one of estimating the mixing proportions of a finite mixture model, with known components. Application of the EM-algorithm to the problem of estimating the mixture proportions of a finite mixture model, where the component densities of the mixture are completely specified, is well-known and relatively straightforward (see for example (McLachlan and Krishnan 1997), section 1.4.3). Note however, that in the case under consideration, each component of the mixture is a *collection* of distributions. Hence, we can write

$$\forall \mathbf{x} \in \mathcal{X} : P(Y|\mathbf{x}) = \sum_{i=1}^{c} \pi_i P_i(Y|\mathbf{x})$$

where $\boldsymbol{\pi} = (\pi_1, \ldots, \pi_c)$ is the vector of unknown mixing proportions, and $\{P_i(Y|\mathbf{x}), \mathbf{x} \in \mathcal{X}\}$ is the collection of distributions, one for each parent configuration $\mathbf{x}$, corresponding to extreme element $i$.

Now, let $\mathcal{D} = \{(y_1, \boldsymbol{x}_1), \ldots, (y_n, \boldsymbol{x}_n)\}$ represent the observed data, and let $z_{ij} = 1$ if $y_j$, with parent configuration $\boldsymbol{x}_j$, originates from component $i$, and $z_{ij} = 0$ otherwise. If $z_{ij}$ were observed, then the MLE of $\pi_i$ would simply be given by (McLachlan and Krishnan 1997)

$$\hat{\pi}_i = \sum_{j=1}^{n} z_{ij}/n \quad (i = 1, \ldots, c),$$

that is, the proportion of the sample originating from the $i$th component of the mixture. The EM scheme to estimate $\pi_i$ is given by (McLachlan and Krishnan 1997)

$$\pi_i^{(t+1)} = \sum_{j=1}^{n} z_{ij}^{(t)}/n$$

for $i = 1, \ldots, c$, where $z_{ij}^{(t)}$ is the posterior probability that $y_j$, with parent configuration $\boldsymbol{x}_j$, originates from the $i$th component of the mixture, based on the current estimate $\boldsymbol{\pi}^{(t)}$ of the mixing proportions, that is

$$z_{ij}^{(t)} = \frac{\pi_i^{(t)} P_i(y_j|\boldsymbol{x}_j)}{\sum_{k=1}^{c} \pi_k^{(t)} P_k(y_j|\boldsymbol{x}_j)},$$

for $i = 1, \ldots, c; j = 1, \ldots, n$.

Finally, we turn to the issue of enumerating the mixture components, or extreme elements. We denote the $N$ distinct parent configurations by $\mathbf{x}^1, \ldots, \mathbf{x}^N$. Recall that each extreme element is a collection of point-mass distributions that satisfies the order constraints. Hence, the set of extreme points $\mathcal{S}$ is given by

$$\mathcal{S} = \{(s_1, \ldots, s_N), s_k \in \mathcal{Y} : \mathbf{x}^i \preceq \mathbf{x}^j \Rightarrow s_i \leq s_j\},$$

where $s_k = y$ indicates that $P(Y|\mathbf{x}^k)$ has point mass on $Y = y$. If we number the elements of $\mathcal{S}$ as $\mathbf{s}_1, \ldots, \mathbf{s}_c$, then we can write

$$P(Y = y|\mathbf{X} = \mathbf{x}^j) = \sum_{k=1}^{c} I(s_{kj} = y)\pi_k, \qquad (3)$$

where $s_{kj}$ denotes the $j$th component of extreme element $\mathbf{s}_k$, and $I(\cdot)$ is the indicator function of the truth of its argument. Hence, after convergence of the EM-algorithm, the final values $\pi_k^{(t)}$ can be plugged into (3) to obtain the required conditional distributions.

Although the mixture approach turns a fairly complex constrained optimization problem into a relative straightforward unconstrained optimization problem, it has the drawback that the number of extreme points grows very fast with the number of parent configurations and the number of elements of $\mathcal{Y}$. If we consider a ternary variable, with ternary parents, then for one parent the number of extreme points is 10, for two parents it's 175 and for three parents it's 211,250. Therefore this algorithm can unfortunately only be applied to small problems. Nevertheless, since it suffices for our purposes, we have used it to perform the experimental comparison with the isotonic regression estimator.

## 5 Experimental Comparison

In this section we report on experiments with the different parameter learning algorithms: the isotonic regression estimator (ISO), the constrained maximum likelihood estimator (CML), and the unconstrained maximum likelihood or standard estimator (STAND). We generated data from a known distribution, computed the different estimators, and compared the fitted distributions with the generating distribution. The similarity between the true and fitted distribution was measured by their Hellinger distance $D_H$:

$$D_H(P, Q) = \sum_{y} \left(\sqrt{P(y)} - \sqrt{Q(y)}\right)^2$$

To generate the data we specified the conditional distribution of $Y$ for each parent configuration, and drew a sample of some fixed size (ranging from 5 to 100) from each of the distributions. The Hellinger distance between the true and fitted distributions was then computed for each conditional distribution and summed.

We first checked whether the standard estimator already satisfied the order constraints, because in that case both order-constrained estimators coincide with the standard estimator. This also allowed us to verify how often there were order reversals in the standard estimator for different distributions and sample sizes.

We first generated data from the distribution depicted in Figure 1. This figure contains all parent configurations for two ternary parents. The arrows indicate the direct precedence relations, imposed by two positive influences, between parent configurations. Below each "layer" of incomparable parent configurations the distribution of the ternary child variable is specified by giving $P(Y \leq 1)$ and $P(Y \leq 2)$. This representation is convenient for verifying whether the order constraints are satisfied. Within each layer, the distributions are identical.

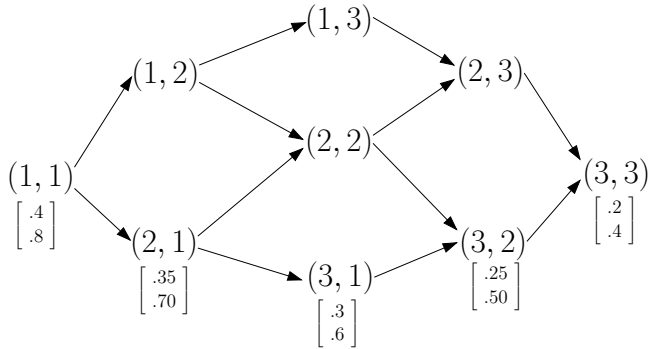

Figure 1: Probability distribution of a network fragment used for data generation.

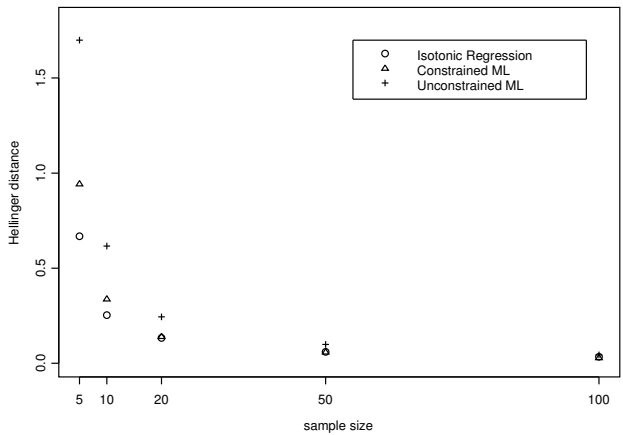

Figure 2: Results for distribution displayed in figure 1.

The results are displayed in Figure 2. This figure shows the Hellinger distance to the true conditional distributions for the different estimators. These distances were added over all nine parent configurations (conditional distributions) and averaged over 100 replications of the experiment. The sample sizes indicated on the x-axis are per parent configuration; for example $n = 5$ means 5 observations per parent configuration, and hence a total sample size of 45 in this case. The figure shows that both constrained estimators perform better than the standard estimator for small sample sizes. As the sample size grows, all three errors become small and almost the same. In this case, it appears the isotonic regression is somewhat better than constrained ML for small sample size, in terms of Hellinger distance.

In order to see what happens when the specified constraints are incorrect, we generated data from the distribution in Figure 1 "reversed", that is the layer 1 distribution was assigned to layer 5, layer 2 was assigned to layer 4, etc. Hence, this corresponds to the rather extreme situation that the expert specified two positive influences, when in fact there are two negative influences.

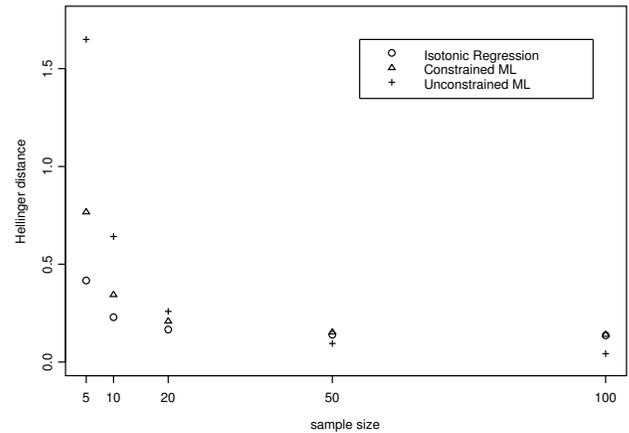

Figure 3: Results for distribution displayed in figure 1 reversed. Constrained estimates were computed with positive influences.

The results are shown in Figure 3. At first sight they seem very surprising: the constrained estimators are still better than the standard estimator, and the error for small sample sizes even appears to be smaller than in the case with correct constraints. The explanation for this is that the constrained estimators benefit from a "smoothing" or "shrinking" effect. This can most clearly be seen for the isotonic regression estimator: order reversals of the standard estimates are removed by averaging them over blocks of "contiguous" parameters; see equation (2). If the constraints applied are exactly the opposite of those actually present in the true distribution, there will be many order reversals in the sample, and hence a lot of averaging to remove these reversals.

In an attempt to correct for this smoothing effect, we applied the Laplace correction for the standard estimator, replaced the constrained ML by the MAP estimator with a prior count of 1 in each cell, and performed the isotonic regression on the Laplace corrected standard estimates. The results are displayed in Figure 4 and Figure 5 respectively.

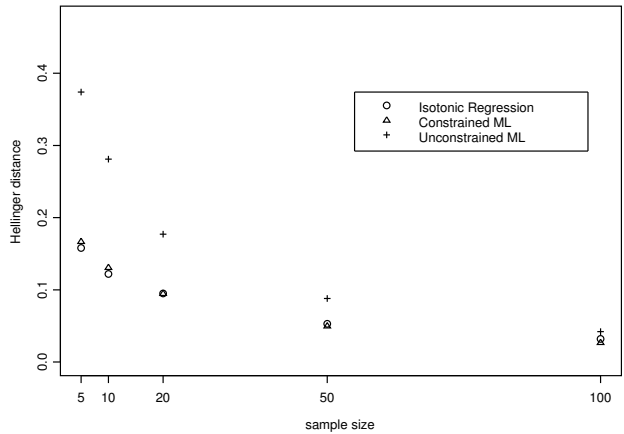

Figure 4: Results for distribution displayed in figure 1, with smoothing (note the different scale on the y-axis).

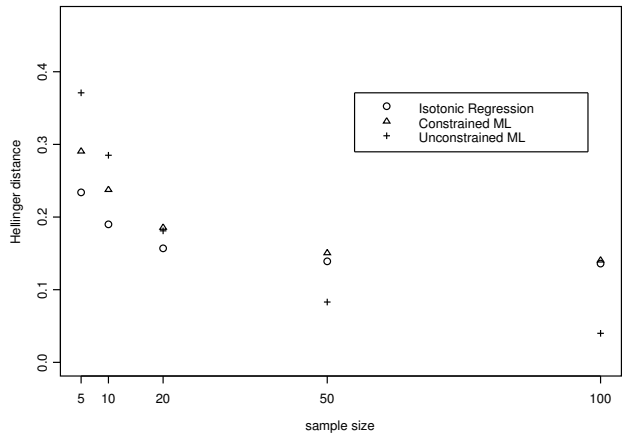

Figure 5: Results for distribution displayed in figure 1 reversed, with smoothing.

We can draw a number of conclusions from these results. Firstly, if we compare Figure 2 with Figure 4, we see that smoothing reduces the error of the estimates for the smaller sample sizes. If we compare Figure 4 with Figure 5, we see that the constrained estimates still give some improvement when the applied constraints are incorrect, but this improvement has now become smaller than when the constraints are correct. With incorrect constraints, the standard estimates already have lower error than constrained ML at $n = 20$, whereas the constrained estimates have somewhat lower error than the standard estimates even at $n = 100$ when the constraints are correct. Nevertheless, Figure 5 shows that for $n = 5, 10$ there is still some benefit from the additional smoothing, on top of the Laplace correction, due to enforcing the incorrect constraints.

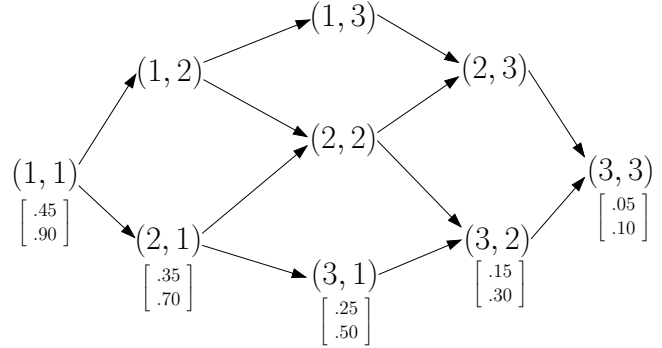

Figure 6: Network fragment with large margin between contiguous parameters used for data generation.

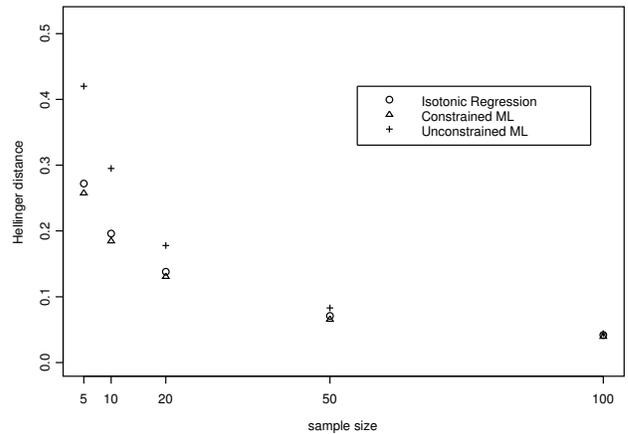

Figure 7: Results for distribution displayed in figure 6, with smoothing.

We performed the same experiments with the probabilities depicted in Figure 6. Here we specified larger margins between the values of contiguous parameters in the ordering. In other words, the positive influences in Figure 6 are stronger than in Figure 1. The results (with smoothing) are depicted in Figure 7 (constraints correct) and Figure 8 (constraints incorrect). In Figure 7 we see that the constrained ML estimator performs slightly better here than the isotonic regression estimator. From Figure 8 we conclude that with the large margin between the values of contiguous parameters, and incorrect constraints, the standard estimator is superior also for small sample size.

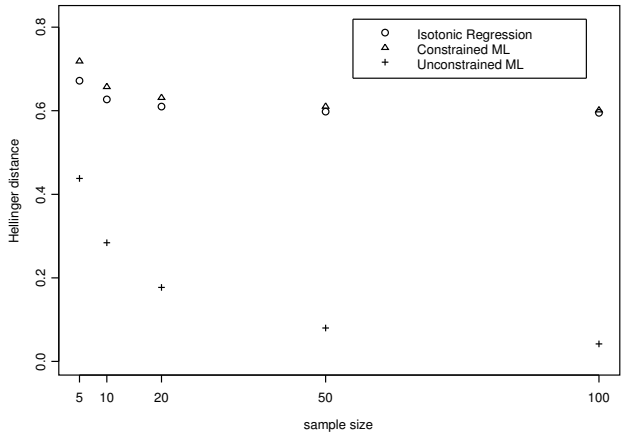

Figure 8: Results for distribution displayed in figure 6 reversed, with smoothing.

In addition to the analysis on simulated data, we also performed a number of experiments on real data sets. We selected two examples where positive influences between parent and child are plausible. Like with the simulated data, we only considered one variable and its parents at a time. The experimental setup was as follows. We drew random samples of size 20, 50 and 100 from the data, and used this sample for estimating the parameters. For the standard estimate we performed the Laplace correction, the isotonic regression was performed on these "smoothed" standard estimates, and the reported constrained ML estimates are actually MAP estimates with a prior count of 1 in each cell. Then we computed the log-loss of each estimator on the remaining data as follows:

$$\mathcal{L}_{\text{test}} = \frac{-\sum_{i=1}^{\text{ntest}} \log \hat{P}(y_i|\boldsymbol{x}_i)}{\text{ntest}},$$

where ntest is the number of observations in the test set. This whole procedure was again replicated 100 times.

The first data set we analyzed is Windsor Housing (Koop 2000). This data set contains the selling price of 546 houses near Windsor (Canada), together with a number of attributes of these houses. For many of these attributes it is plausible that they have a positive relation with the selling price. We selected the lot size and number of bedrooms as parents of selling price. The selling price was discretized into four levels, using the quartiles of the empirical distribution. Parent variables were subsequently discretized with the algorithm of Fayyad and Irani (1993). This resulted in three levels for lot size, and two levels for number of bedrooms.

| $n$ | ISO | CML | STAND | % REV |
|---|---|---|---|---|
| 20 | 1.249 | 1.248 | 1.273 | 97% |
|    | ± .029 | ± .030 | ± .042 | |
| 50 | 1.215 | 1.214 | 1.224 | 92% |
|    | ± .023 | ± .023 | ± .027 | |
| 100 | 1.192 | 1.192 | 1.196 | 81% |
|    | ± .019 | ± .019 | ± .021 | |

Table 1: Average log-loss ± standard error, of the different estimators on Windsor housing data. The final column shows how many times out of the 100 replications there were order reversals in the standard estimates.

| $n$ | ISO | ML | STAND | % REV |
|---|---|---|---|---|
| 20 | 1.112 | 1.115 | 1.154 | 100% |
|    | ± .022 | ± .024 | ± .040 | |
| 50 | 1.109 | 1.111 | 1.140 | 100% |
|    | ± .023 | ± .023 | ± .027 | |
| 100 | 1.096 | 1.097 | 1.112 | 100% |
|    | ± .017 | ± .018 | ± .024 | |

Table 2: Average log-loss ± standard error, of the different estimators on Pima Indians data.

The results are presented in Table 1. We read from the table that constrained maximum likelihood (CML) and isotonic regression (ISO) perform somewhat better than the standard estimator (STAND) on the test data for small training samples. We conclude from the reported standard errors however, that the observed improvements are not significant. The performance of CML and ISO is seen to be virtually identical.

The same basic observations hold for a second example, where we use the Pima Indians data from the UCI machine learning repository (Blake and Merz 1998). The example is a fragment from the network specified by Altendorf, Restificar, and Dietterich (2005). The body-mass index is the child variable, and age, number of pregnancies and perceived risk due to pedigree are its parents, all with a positive influence. The body-mass index was discretized into three equal-frequency bins, and the parents were again discretized using the algorithm of Fayyad and Irani (1993). This resulted in three binary parents. The results are reported in Table 2.

Note that even for $n = 100$, the standard estimates have order reversals on all training samples. Computation of the standard estimates reveals that one of the specified order constraints (the positive influence of number of pregnancies) is violated on the complete data set.

## 6  Summary and Conclusions

We have presented a new method for learning the parameters of Bayesian networks with prior knowledge of qualitative influences. We experimentally compared its performance to that of constrained maximum likelihood estimation and standard (unconstrained) estimation. The results on simulated data show that our isotonic regression estimator has performance comparable to that of the constrained ML estimator, and that both perform better than the standard estimator for relatively small sample sizes. A similar observation holds for the results on real data, although the performance improvement, in terms of log-loss, could not be shown to be significant.

The constrained maximum likelihood estimates are fairly complicated to compute, whereas computation of the isotonic regression estimates only requires the repeated application of the PAV algorithm for linear orders. Therefore, the isotonic regression estimator is to be preferred from the viewpoint of computational complexity.